\documentclass[twoside,11pt]{article}

\usepackage{blindtext}
\usepackage[abbrvbib,preprint]{jmlr2e}

\usepackage{amssymb}
\usepackage{booktabs}
\usepackage{enumitem}
\usepackage{longtable}
\usepackage{graphicx}
\usepackage{array}
\usepackage{dcolumn}
\usepackage{amsfonts}
\usepackage{amssymb}
\usepackage{xspace}
\usepackage{xfrac}
\usepackage{array}
\usepackage{multirow}
\usepackage{xcolor}
\usepackage{footnote}
\usepackage{url}
\usepackage{makeidx}
\usepackage{epsfig}
\usepackage{amsmath}
\usepackage{multicol}
\usepackage{multirow}
\usepackage{mathtools}
\usepackage[misc]{ifsym}
\usepackage[ruled,linesnumbered]{algorithm2e}
\usepackage{tabularx}
\usepackage{array}
\usepackage{footnote}
\makesavenoteenv{tabular}
\makesavenoteenv{table}
\usepackage{subcaption}

\firstpageno{1}

\begin{document}

\title{MetaARIMA: Configuration-Level Transfer Learning for ARIMA Models}

\author{\name Vitor Cerqueira 
        \email vitorc@dei.uc.pt \\
       \addr University of Coimbra\\
       CISUC/LASI – Centre for Informatics and Systems of the University of Coimbra\\
       Coimbra, Portugal
       \AND
       \name Ricardo Inácio 
       \email rcinacio@fe.up.pt \\
       \addr Faculty of Engineering, University of Porto\\ Porto, Portugal
       \AND
       \name Carlos Soares
       \email csoares@fe.up.pt \\
       \addr Faculty of Engineering, University of Porto\\ 
       Laboratory for Artificial Intelligence and Computer Science (LIACC), Portugal\\
        Fraunhofer Portugal AICOS, Portugal\\
       Porto, Portugal
       }

\editor{My editor}

\maketitle

\begin{abstract}

Time Series Foundation Models have demonstrated that pretraining on diverse datasets enables effective transfer to new forecasting tasks. Classical methods like ARIMA, however, remain configured and trained independently for each series, discarding knowledge from previous experiments. Consequently, selecting and training an ARIMA configuration remains challenging and often relies on heuristic approaches based on in-sample goodness-of-fit measures. 
We propose \texttt{MetaARIMA}, a pretrained configuration policy based on metalearning that brings transfer learning to classical statistical forecasting at the configuration level rather than the model level. The policy learns from historical experiments which ARIMA configurations achieve top-quantile performance, and recommends promising candidates for new series while preserving interpretability and local estimation. At inference, it selects a diverse set of candidates via maximal marginal relevance and efficiently identifies the final configuration using successive halving. Experiments in a transfer learning setting, training on M4 and evaluating on M1, M3, and Tourism, show that \texttt{MetaARIMA} outperforms AutoARIMA and other classical methods while being more data-efficient. In a comparison with recent zero-shot foundation models, \texttt{MetaARIMA} remains competitive with Chronos2, Moirai2, and TimesFM. The experiments are publicly available at \url{https://github.com/vcerqueira/experiments-metaarima}.

\end{abstract}


\section{Introduction}

{Recent advances in time series forecasting have been driven by pretrained models that learn transferable representations from large collections of time series. Time Series Foundation Models (TSFMs) such as Chronos~\citep{ansari2024chronos}, TimesFM~\citep{das2024decoder}, and Moirai~\citep{liu2025moirai} demonstrate that knowledge acquired during pretraining on diverse datasets can transfer effectively to new forecasting tasks, achieving strong zero-shot performance without task-specific fine-tuning. This transfer learning paradigm, where a model learns general patterns from historical data and applies them to unseen problems, has become a central theme in modern forecasting research.

Classical statistical methods such as ARIMA (Auto-Regressive Integrated Moving-Average)~\citep{box2015time} have not benefited from this transfer learning revolution. Despite their widespread use, interpretability, and theoretical grounding, ARIMA models are configured independently for each new time series. Automated methods like AutoARIMA~\citep{hyndman2008automatic} rely on heuristic search guided by in-sample fit criteria such as AICc, while manual approaches like the Box-Jenkins method~\citep{makridakis1997arma} require significant expertise. Crucially, both approaches are \emph{memoryless}: they discard all knowledge from previous configuration experiences, reconfiguring each new series from scratch.

This paper introduces \texttt{MetaARIMA}, a \emph{pretrained configuration policy} for ARIMA models that brings the transfer learning paradigm to classical statistical forecasting.\footnote{Throughout this paper, an ARIMA \emph{configuration} refers to the specification of the model orders $(p, d, q)$ and, for seasonal series, $(P, D, Q)$.} Just as TSFMs pretrain a forecasting model on diverse time series and apply it to new tasks, \texttt{MetaARIMA} pretrains a configuration policy on historical experiments and applies it to configure ARIMA for new series. The key insight is that transfer learning can operate at the configuration level rather than the model level: instead of pretraining the forecaster itself, we pretrain a lightweight controller that selects appropriate ARIMA specifications based on time series characteristics. This preserves the interpretability and local estimation properties of ARIMA while enabling it to benefit from accumulated historical experiments.}

{We frame ARIMA configuration as an algorithm selection problem~\citep{rice1976algorithm} and employ metalearning~\citep{brazdil2008metalearning,moniz2021automated} to train the configuration policy.} The policy is trained on a meta-dataset of (time series features, configuration performance) pairs collected from a large source collection. Rather than predicting a single best configuration, it identifies which configurations are likely to fall within a top quantile of performance for a given series. {This design choice addresses the inherent uncertainty in configuration selection: performance differences among top candidates are often marginal, so predicting a diverse set of promising options is more robust than committing to a single choice which is a common approach in algorithm selection tasks~\citep{brazdil2008metalearning}.}

{Technically, the configuration policy works as follows.} Because the ARIMA configuration space can be large (e.g., hundreds of candidate orders for a given time series), directly modelling top-quantile membership for every configuration yields a high-dimensional output space. {To reduce this dimensionality, we apply} principal component analysis (PCA) to the binary top-quantile indicators, obtaining a compact representation that captures dominant patterns of configuration performance. The resulting problem becomes a multi-target regression from time series features to principal components, whose predictions can be mapped back to configuration space to recover the score of each configuration belonging to the top quantile.

At inference, {the pretrained policy recommends} a subset of $k$ promising candidates {for a new time series}. Candidate selection balances two criteria: (i) the predicted configuration score reflecting top-quantile affinity, and (ii) maximal marginal relevance (MMR)~\citep{cerqueira2019arbitrage} to promote diversity. Because configurations are often highly correlated in performance, encouraging diversity reduces redundancy and improves the likelihood that at least one candidate performs well. From these $k$ candidates, we select the final configuration using successive halving~\cite{li2018hyperband}: configurations are evaluated on small data subsets first, the worst half is discarded, and the process repeats on progressively larger subsets until one configuration remains. 
{This two-stage process, which consists of a pretrained policy for candidate generation followed by efficient local evaluation on the target series, mirrors the pretrain-and-adapt paradigm common in foundation model applications~\citep{liang2024foundation}.}

We evaluate \texttt{MetaARIMA} in a transfer learning setting where the {configuration policy} is trained on the M4 dataset~\citep{makridakis2018m4} and evaluated on standard forecasting benchmarks: M1~\citep{makridakis1982m1}, M3~\citep{makridakis2000m3}, and Tourism~\citep{athanasopoulos2011tourism}. Separate {policies} are trained for each sampling frequency (monthly, quarterly, yearly) and applied to the corresponding test datasets. The results demonstrate that \texttt{MetaARIMA} consistently outperforms AutoARIMA, the state-of-the-art method for configuring ARIMA models, as well as other classical forecasting methods such as ETS and Theta. We also compare \texttt{MetaARIMA} with recent TSFMs in a zero-shot setting, finding that it performs competitively with Chronos2, Moirai2, and TimesFM, while retaining the inherent interpretability of ARIMA.  The experiments are publicly available at \url{https://github.com/vcerqueira/experiments-metaarima}.

The main contributions of this work are:
\begin{itemize}
    \item {We propose a new paradigm for classical forecasting: transfer learning at the \emph{configuration} level rather than the model level. \texttt{MetaARIMA} is a pretrained configuration policy, a lightweight controller that maps time series characteristics to promising ARIMA specifications, which enables classical models to benefit from accumulated configuration knowledge while preserving local estimation and interpretability.}
    \item {We position this paradigm between conventional memoryless ARIMA configuration and pretrained neural forecasters: \texttt{MetaARIMA} transfers knowledge across series like TSFMs, but adapts a local statistical model rather than replacing it with a global neural forecaster.}
    \item {Empirically, we show that this configuration-level transfer is effective: \texttt{MetaARIMA} outperforms AutoARIMA and classical baselines across diverse benchmarks and remains competitive with state-of-the-art TSFMs.}
\end{itemize}

The rest of this paper is organized as follows. Section~\ref{sec:background} provides a background on time series forecasting, ARIMA models, and metalearning. Section~\ref{sec:metaarima} describes the proposed \texttt{MetaARIMA} approach. The experimental setup and results are presented in Section~\ref{sec:experiments}. Section~\ref{sec:discussion} discusses the findings and limitations. Finally, Section~\ref{sec:conclusions} concludes the paper and outlines future work directions.

\section{Background}\label{sec:background}

\subsection{Univariate Time Series Forecasting}

A time series $Y$ of length $t$ is defined as $\{y_1, y_2, \dots,$ $y_t \}$, where $y_i \in \mathbb{R}$ is the $i$-th value of $Y$. We consider univariate time series, where a single real-valued variable is observed at each time step.  

Forecasting denotes the process of predicting the next $h$ values of $Y$, $y_{t+1}, \ldots, y_{t+h}$, based on historical data, where $h$ is the forecasting horizon. 
Forecasting problems often involve time series collections that contain multiple univariate time series. We define a time series collection as $\mathcal{Y} = \{Y_1, Y_2, \dots, Y_N\}$, where $N$ is the number of time series in the collection. 

Forecasting methods can be broadly categorized as either local or global approaches~\cite{januschowski2020criteria}. Local methods, such as ARIMA and exponential smoothing, involve fitting a separate model to each time series in a collection $\mathcal{Y}$. In contrast, global methods (typically based on machine learning techniques like neural networks or tree-based algorithms) train a single model on data pooled from many time series, leveraging shared patterns across the collection. 

Although ARIMA models are typically fitted using the local approach, our work introduces a global trait to the configuration process: we employ a metalearning framework that leverages historical performance across multiple series to guide configuration for new time series. Thus, while the final ARIMA models remain locally fitted, our configuration procedure incorporates global information, resulting in a hybrid strategy that combines the strengths of both local and global approaches.

\subsection{ARIMA}

ARIMA models combine autoregressive terms, moving-average terms, and differencing to capture temporal dependencies and handle non-stationarity~\citep{box2015time}. A non-seasonal ARIMA model is specified by three orders $(p, d, q)$: the number of autoregressive lags, the degree of differencing, and the number of moving-average lags. For seasonal data, the model extends to ARIMA$(p,d,q)(P,D,Q)_m$, where $(P, D, Q)$ are seasonal counterparts applied at lag $m$ (the seasonal period). We refer the reader to~\cite{box2015time} for a comprehensive read on ARIMA models.


ARIMA models can be configured in a variety of ways by choosing the specific orders for their components $(p, d, q, P, D, Q)$. Selecting the most appropriate configuration is crucial to getting the most performance out of ARIMA. This process poses several challenges, and multiple approaches have been proposed in the literature.

One of the most influential techniques for ARIMA configuration is the Box-Jenkins methodology~\citep{box2015time}, which is a systematic, iterative manual procedure consisting of three phases: identification, estimation, and diagnostic checking. During identification, the practitioner carefully examines the autocorrelation function (ACF) and partial autocorrelation function (PACF) plots to determine suitable values for the AR and MA orders, as well as the necessary degree of differencing to achieve stationarity. Once a candidate model is selected, parameters are estimated, and residuals are analyzed for white noise; if the diagnostics are unsatisfactory, the process cycles back to explore alternative orders. Although this methodology is robust and well-established, it demands significant expertise and manual effort, making it impractical for configuring ARIMA models across large collections of time series.

To address these limitations, automated procedures such as AutoARIMA~\citep{hyndman2008automatic} have been developed. AutoARIMA streamlines the configuration process by first applying unit-root tests (such as KPSS) to determine the order of differencing and then performing a stepwise search over AR and MA orders. The selection process is guided by an information criterion, typically AICc, which seeks to balance model fit and complexity. While AutoARIMA eliminates the need for manual inspection and is scalable to many time series, it relies on in-sample fit and a heuristically defined search path, which can result in suboptimal configurations.

Another method for configuring ARIMA, and algorithm selection in general, is to directly assess a set of candidate configurations based on out-of-sample performance. This strategy evaluates model performance on held-out data, providing out-of-sample error estimates that are generally more reflective of future accuracy than in-sample criteria. However, exhaustively exploring a large configuration space in this manner can be computationally prohibitive, particularly when dealing with a substantial number of time series.

A common limitation of all these approaches is that they are \emph{memoryless}: they do not leverage information from configuring ARIMA on previous time series. {We return to this limitation in Section~\ref{sec:paradigms}, where we contrast local configuration methods with global pretrained forecasters and introduce configuration-level transfer learning as a middle ground.}

\subsection{Time Series Feature Extraction}\label{sec:feature-ext}

Feature extraction is a crucial step for systematically characterizing time series. Because time series can vary in length and structure, directly comparing them is often challenging. By mapping each series to a common set of descriptive statistics or attributes, we achieve dimensionality reduction and enable robust, meaningful comparisons across diverse series (e.g.~\citep{cerqueira2024vest}). These features can capture a wide range of properties, such as trend, seasonality, or spectral characteristics, providing a standard representation suited to both visualization and downstream tasks.

Several automated frameworks have been developed to extract such features from time series. Notable examples include \texttt{tsfeatures}~\citep{tsfeatures}, \texttt{tsfel}~\citep{barandas2020tsfel}, and \texttt{tsfresh}~\citep{christ2018time}. Among these, \texttt{tsfeatures} is especially popular in forecasting applications, such as weighted forecast combination~\citep{montero2020fforma}. It provides a set of 42 interpretable features, encompassing attributes such as linearity, lumpiness, and seasonal strength, that are widely used in both academic and industrial settings.

\subsection{Metalearning for Algorithm Selection}


Metalearning denotes the process of using historical experiments to improve the learning process of algorithms~\citep{brazdil2008metalearning,moniz2021automated}. In the context of algorithm selection, the goal is to choose, for a given problem (e.g.~a time series), an algorithm (or configuration) that is expected to perform well, rather than evaluating all options from scratch. The algorithm selection problem was formalized by Rice~\cite{rice1976algorithm}: problem instances are described by a set of features, and a selection model maps these features to a recommended algorithm. The model is typically learned from a meta-dataset of (instance features, algorithm performance) pairs. Thus, metalearning for algorithm selection exploits historical performance to predict which algorithm suits a new instance, reducing the need for exhaustive evaluation.


Metalearning has been successfully applied in time series forecasting in several ways. In \emph{forecast combination}, for example, meta-models use time series features to assign weights to base forecasters (e.g.,~\citep{montero2020fforma}), improving accuracy by leveraging the relative strengths of different methods across series. 
In \emph{dynamic model selection and weighting}, meta-models forecast the future errors of candidate forecasting models, and these error predictions are then used to dynamically weight each model's prediction at every time step~\citep{cerqueira2019arbitrage}. Another application is multi-step ahead forecasting~\citep{cerqueira2024instance}, where metalearning is used to introduce a conditional dependency constraint across different forecasting approaches, such as recursive or direct methods.

Algorithm selection has also been studied in time series (e.g., choosing among forecasting methods or hyperparameters based on meta-features) (e.g.~\cite{lemke2010meta,santos2025metafore}). Despite these advances, using metalearning to \emph{configure} ARIMA, i.e., to select (p, d, q, P, D, Q) from historical configuration performance, has not been explored. We fill this gap by framing ARIMA configuration as an algorithm selection problem and training a meta-model that maps time series features to promising ARIMA configurations.

{\subsection{Research Gap and Positioning}\label{sec:paradigms}

Recent advances in global forecasting methods and TSFMs have expanded the design space for time series forecasting, where different paradigms occupy distinct positions along the global-local trade-off spectrum~\citep{januschowski2020criteria}.

Classical methods such as ARIMA and exponential smoothing fit a separate model to each time series. Configuration procedures like AutoARIMA~\citep{hyndman2008automatic} and the Box-Jenkins methodology~\citep{box2015time} operate independently on each series, using in-sample criteria or manual inspection to select model orders. These approaches are \emph{memoryless}: they do not retain or transfer any knowledge from configuring previous time series. Each new series is treated as an isolated problem, discarding potentially valuable experience from historical experiments.

At the other end of the spectrum, TSFMs pretrain a single neural forecaster on large, diverse time series collections and apply it directly to new series~\citep{liang2024foundation}. These models also transfer learned representations across downstream tasks such as forecasting, anomaly detection, and missing value imputation~\citep{liang2024foundation}. However, they replace the local statistical model entirely with a global neural architecture, sacrificing the interpretability, parsimony, and theoretical grounding of classical methods. We refer the reader to~\citep{liang2024foundation} for a comprehensive review on TSFMs.

We propose a middle ground: transfer learning at the \emph{configuration} level rather than the model level. Instead of pretraining the forecaster itself like TSFMs, we pretrain a \emph{configuration policy} that maps time series features to promising ARIMA specifications. This policy learns from historical configuration experiments across many series and generalizes to new ones, akin to the transfer learning principle of TSFMs, but the final forecasting model remains a locally fitted, interpretable ARIMA. This paradigm combines the knowledge transfer benefits of global approaches with the transparency and local estimation properties of classical methods.

\texttt{MetaARIMA} instantiates a configuration-level transfer learning paradigm: it pretrains a configuration policy on the M4 dataset and applies it to recommend ARIMA specifications for new time series, enabling classical models to benefit from accumulated configuration historical data while preserving their inherent interpretability.}

\section{\texttt{MetaARIMA}}\label{sec:metaarima}

This section describes \texttt{MetaARIMA}, a pretrained configuration policy that implements the configuration-level transfer learning paradigm introduced in Section~\ref{sec:paradigms}. The policy is trained once on a large source collection and then applied, without retraining, to recommend ARIMA configurations for new time series. We first formulate the problem and outline the design rationale that motivates each component (Section~\ref{sec:rationale}), then detail the development stage in which the policy is trained (Section~\ref{sec:metaarima1}) and the inference stage in which it is applied (Section~\ref{sec:metaarima2}).

\subsection{Problem Formulation}\label{sec:rationale}

{For a new time series $Y_{query}$, our goal is to select an ARIMA configuration $(p, d, q, P, D, Q)$ that is expected to yield the best possible forecast accuracy. The configuration space $\mathcal{A}$ can be large (e.g., 400 candidates for seasonal series), exhaustive evaluation is expensive, and in-sample criteria such as AICc may not reflect out-of-sample accuracy. Our goal is to leverage historical configuration experiments from a source collection $\mathcal{Y}$ to guide configuration for $Y_{query}$ efficiently and effectively.

Building an ARIMA configuration policy that transfers across a collection of time series involves four key challenges related to the algorithm selection problem:

\begin{enumerate}
    \item Marginal differences among top configurations: For a given series, many ARIMA configurations achieve similar forecasting accuracy, and the single best is often indistinguishable from several near-best alternatives. In effect, the common approach in algorithm selection of predicting a single winner is therefore unreliable. We frame the target for metalearning as top-quantile membership rather than single-best prediction, identifying the set of configurations that perform well rather than committing to a single one as often done in algorithm selection~\cite{brazdil2008metalearning}.
    
    \item High-dimensional output space: With hundreds or more possible ARIMA configurations, especially for seasonal series, directly modelling top-quantile membership yields a high-dimensional multi-label target. We apply PCA to compress the binary top-quantile indicators into a low-dimensional representation that captures dominant patterns of configuration performance, then train a multi-target regressor on the compressed targets.
    
    \item Redundancy among configurations: Similar ARIMA orders tend to succeed or fail together, so a shortlist of top-scoring configurations may contain significant redundancy. At inference, we use maximal marginal relevance (MMR) to balance predicted performance with diversity to ensure that the candidate set covers configurations that are promising performance-wise and not too similar to each other.
    
    \item Efficient final selection: Evaluating all shortlisted candidates on the full training data is still costly, especially for long time series. We apply successive halving, progressively discarding the worst-performing configurations on increasingly large data subsets, so that computational effort concentrates on promising candidates.
\end{enumerate}

Overall, \texttt{MetaARIMA} addresses these challenges by framing the configuration selection problem as a metalearning task and training a configuration policy that maps time series features to promising ARIMA configurations. The remainder of this section details each stage.}

\subsection{Development stage}\label{sec:metaarima1}

\begin{figure}[ht]
    \centering
    \includegraphics[width=\textwidth]{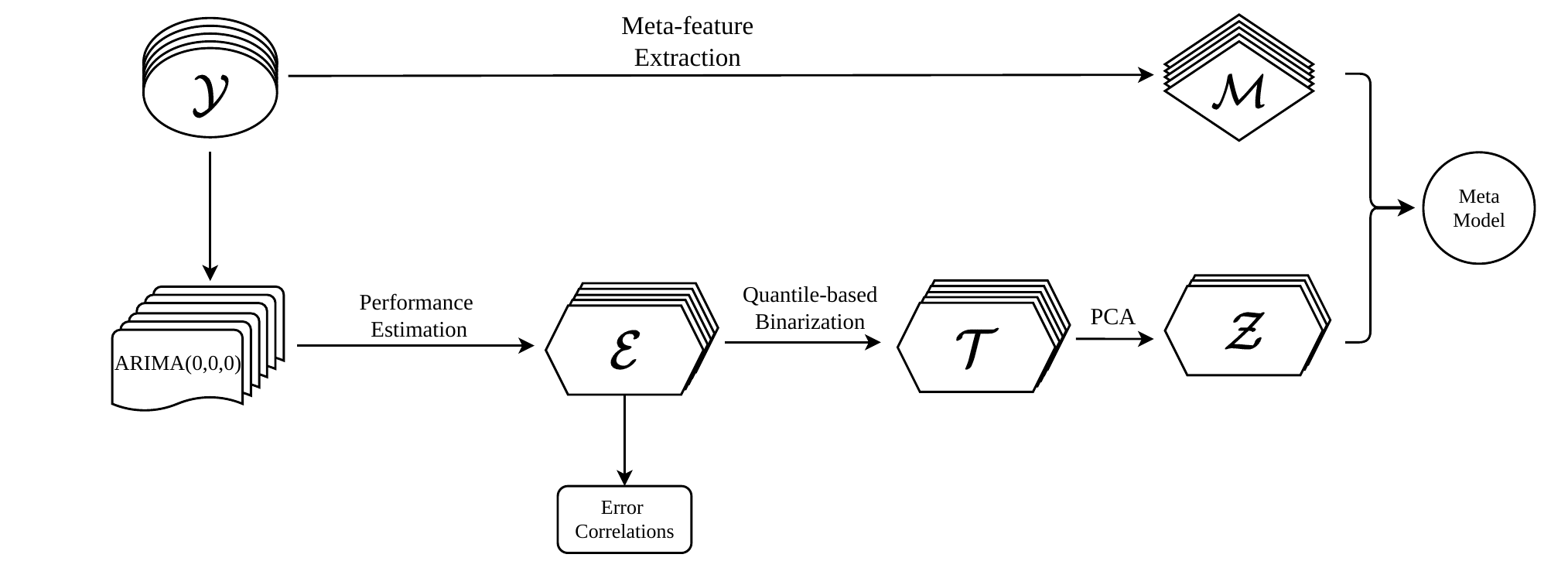}
    \caption{Workflow for building \texttt{MetaARIMA}. A set of time series $\mathcal{Y}$ is summarised into a feature set $\mathcal{M}$. A set of ARIMA configurations (e.g.\ ARIMA(0,0,0)) is evaluated on each series in $\mathcal{Y}$, resulting in error estimates $\mathcal{E}$. The error scores $\mathcal{E}$ are processed according to a quantile-based binarization that identifies series-wise top configurations ($\mathcal{T}$), a representation that is then compacted using PCA ($\mathcal{Z}$). A correlation analysis is also conducted using $\mathcal{E}$ to identify redundant configurations.}
    \label{fig:metaarima1}
\end{figure}

Figure~\ref{fig:metaarima1} outlines the development workflow that trains the configuration policy. {Starting from a source collection $\mathcal{Y}$, we (i) extract meta-features $\mathcal{M}$ that characterize each series, (ii) evaluate all ARIMA configurations on each series to obtain out-of-sample error estimates $\mathcal{E}$, (iii) compute a correlation matrix $R$ to capture redundancy among configurations, (iv) binarize $\mathcal{E}$ into top-quantile indicators $\mathcal{T}$ and compress them via PCA to $\mathcal{Z}$, and (v) train a multi-target regressor to predict $\mathcal{Z}$ from $\mathcal{M}$. The following subsections detail each step.}

\subsubsection{Performance estimation}

Let $\mathcal{A} = \{A_1, \dots, A_a\}$ denote the set of ARIMA configurations. For each time series $Y \in \mathcal{Y}$, we split it into training and test sets, with the test set comprising the last $h$ observations (the forecasting horizon). We fit each configuration in $\mathcal{A}$ on the training set and evaluate its forecasts on the test set using a forecasting error metric such as mean absolute scaled error (MASE). Unlike AutoARIMA, which relies on in-sample goodness-of-fit, we use these out-of-sample error estimates to drive configuration selection, as they better reflect predictive performance on unseen data~\citep{tashman2000out}.

The procedure yields, for each series $i$, a vector $E_i = (e^1_i, \dots, e^a_i)$ of error scores, where $e^j_i$ is the error of configuration $A_j$ on series $Y_i$. We collect these vectors in $\mathcal{E} = \{E_1, \dots, E_N\}$, which will form the targets for the meta-model. We also compute the $a \times a$ correlation matrix $R$ from $\mathcal{E}$: $R_{ij}$ is the Pearson correlation between the error vectors of configurations $A_i$ and $A_j$ across the $N$ series. High $R_{ij}$ indicates that the two configurations tend to succeed or fail on the same series (redundancy). At inference, $R$ is used to promote diversity by favoring configurations with low correlation to those already selected.

\subsubsection{Meta-feature extraction}

The meta-model takes as input a fixed-length vector of numerical features for each time series. Because series can differ in length and scale, feature extraction maps each $Y \in \mathcal{Y}$ to a common representation so that different series can be compared and fed to the same model. 

Formally, we define a feature extraction function $g$ that maps each time series $Y_i$ to a vector $M_i = g(Y_i) \in \mathbb{R}^l$, where $l$ is the number of features. The set of meta-features is $\mathcal{M} = \{M_1, \dots, M_N\}$, which serves as the input to the meta-model alongside the processed targets derived from $\mathcal{E}$.

Features typically capture statistical and structural properties of the series, such as measures of trend, seasonality, stationarity, autocorrelation, or entropy~\citep{fulcher2018feature}. In the experiments, we use a standard extractor (e.g., \texttt{tsfeatures}) that produces $l$ such descriptors per series.

{\subsubsection{Quantile-based target representation and dimensionality reduction}}

{Recall from Section~\ref{sec:rationale} that many ARIMA configurations often exhibit similar forecasting accuracy for a given series, making single-best prediction an unreliable approach (Challenge~1), and that the configuration space can contain hundreds or more candidates, yielding a high-dimensional target that is computationally prohibitive to model directly (Challenge~2). We address both issues here by (i)~framing the target as top-quantile membership and (ii)~compressing that representation with PCA.}

The meta-model is trained on $\mathcal{M}$ and $\mathcal{E}$. Rather than predicting raw error scores (which are scale-dependent and make the identity of the best configuration unstable), we turn $\mathcal{E}$ into binary indicators of whether each configuration lies in the top fraction of performance for that series. Let $\tau \in (0,1)$ be the chosen quantile level (e.g., top 25\%). For each series $Y_i$, let $\tau_i$ be the $\tau$-th quantile of its error vector $(e^1_i, \dots, e^a_i)$. We define the binary target $t^j_i$ for configuration $A_j$ on series $Y_i$ as
\begin{equation}
    t^j_i = \begin{cases}
        1 & \text{if } e^j_i \leq \tau_i \\
        0 & \text{otherwise}
    \end{cases}
\end{equation}
\noindent and collect these in vectors $T_i = (t^1_i, \dots, t^a_i)$ and the set $\mathcal{T} = \{T_1, \dots, T_N\}$. Thus, $T_i$ identifies which configurations achieved top-quantile performance on $Y_i$; multiple configurations can be 1 for the same series (multi-label structure, cf.~\citep{read2011classifier}). This quantile-based binarization differs from the standard algorithm selection formulation, which typically predicts raw performance scores or a single best configuration; framing the target as multi-label top-quantile membership better captures the fact that many ARIMA configurations perform comparably well on a given series.

Modeling $\mathcal{T}$ directly would require predicting $a$ binary labels per series (one for each configuration). Classifier chains~\citep{read2011classifier} solve this task by chaining one binary classifier per label, which would mean training $a$ models, which, in the case of the ARIMA configurations, would be hundreds. Beyond the computational cost, many of these binary targets are highly correlated because similar ARIMA orders tend to succeed or fail together, so modeling each label independently wastes modeling capacity on redundant variables.

To address this problem, we reduce the dimension of $\mathcal{T}$ with PCA, obtaining a set of principal components $\mathcal{Z}$ of length $n_{pc} \ll a$, and fit a single multi-target regressor that predicts $\mathcal{Z}$ from $\mathcal{M}$. This dimensionality reduction in the label space is an instance of Principal Label Space Transformation~\citep{tai2012multilabel}, a technique established in multi-label learning that, to our knowledge, has not previously been applied to algorithm selection or forecasting. PCA captures the dominant patterns of configuration performance in a compact representation, effectively acting as a regularizer that discards noisy, low-variance variation. This compression also makes the trained meta-model lightweight. Formally, the training data are input-output pairs $(\mathcal{M}, \mathcal{Z})$, and we fit a model $f: \mathbb{R}^l \rightarrow \mathbb{R}^{n_{pc}}$, where $l$ is the number of meta-features and $n_{pc}$ the number of retained components. At inference, given features $M_{query}$ for a new series, we predict $\hat{Z} = f(M_{query})$, then apply the inverse PCA transform to obtain a score  for each configuration, which we use to rank and select candidates.

\subsection{Inference stage}\label{sec:metaarima2}

\begin{figure}[ht]
    \centering
    \includegraphics[width=\textwidth]{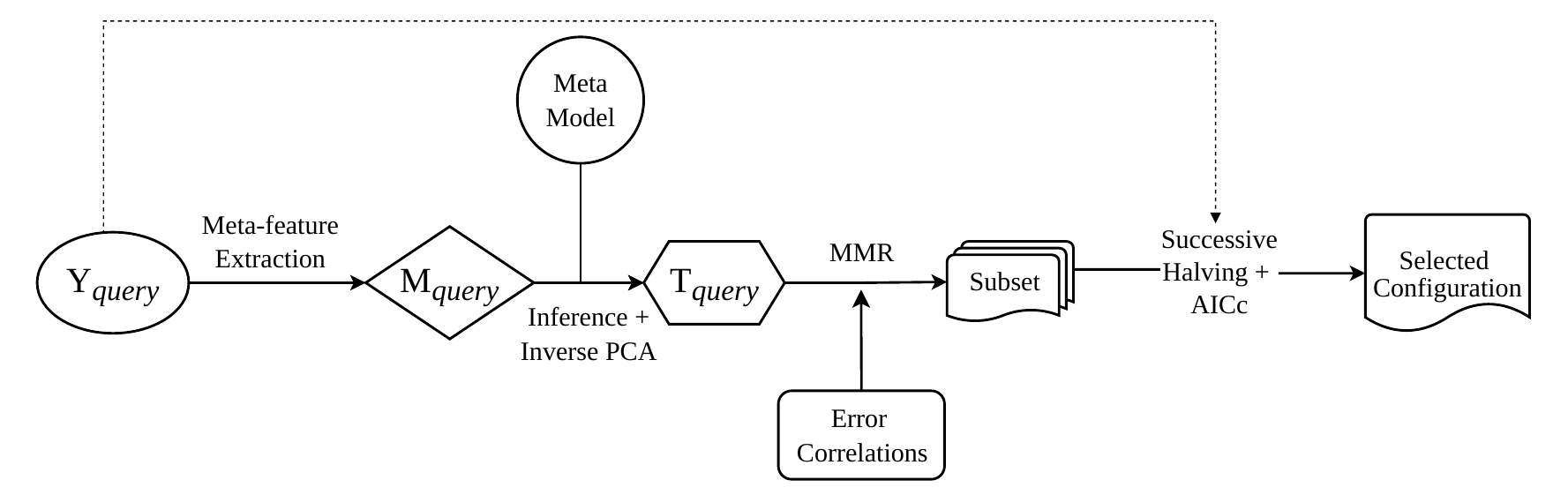}
    \caption{Workflow for the inference stage of \texttt{MetaARIMA}. The query time series $Y_{query}$ is converted to meta-features $M_{query}$; the meta-model outputs predictions in PCA space, which are inverted to obtain scores $\hat{T}_{query}$. MMR, using $\hat{T}_{query}$ and the stored error correlations, selects a diverse subset of configurations; successive halving and AICc on $Y_{query}$ then choose the single selected configuration.}
    \label{fig:metaarima2}
\end{figure}

Figure~\ref{fig:metaarima2} outlines the inference workflow for a new query time series $Y_{query}$. We first extract meta-features $M_{query} = g(Y_{query})$. The meta-model predicts in the PCA space $\hat{Z}_{query} = f(M_{query})$, and we apply the inverse PCA transform to obtain $\hat{T}_{query}$, a vector of scores  for each configuration $A_j \in \mathcal{A}$, which we interpret as estimates of $\mathbb{P}(\text{configuration\ $A_j$ in top quantile}\mid M_{query})$. 

Using $\hat{T}_{query}$ and the precomputed correlation matrix $R$, we select a subset of $k$ promising, diverse configurations. From these, we choose the final configuration by evaluating them on $Y_{query}$ with successive halving and AICc. The next two subsections detail the subset selection (top-$k$ via MMR) and the successive-halving step.

{\subsubsection{Diverse candidate selection via MMR}}

{As noted in Section~\ref{sec:rationale}, similar ARIMA orders tend to succeed or fail together, so simply selecting the $k$ highest-scoring configurations could yield a redundant shortlist (Challenge~3). We therefore balance predicted performance with diversity using MMR~\citep{carbonell1998use}, a ranking criterion originally developed for information retrieval and previously adapted to forecast combination~\citep{cerqueira2019arbitrage}. The gist of using MMR is that each new candidate is chosen based on its predicted score while remaining weakly correlated with those previously selected (cf.~\citep{carbonell1998use}).} 

The first configuration is the one with the highest entry in $\hat{T}_{query}$. Each subsequent configuration is chosen from the remaining candidates by maximizing the MMR score: for candidate $c$ and current set $S$ of already selected configurations, we set
\begin{equation}
    \mathrm{MMR}(c) = \lambda \cdot \hat{t}_c - (1-\lambda) \cdot \max_{s \in S} R_{s,c},
\end{equation}
\noindent where $\hat{t}_c$ is the predicted score for configuration $c$ (the $c$-th entry of $\hat{T}_{query}$), $R_{s,c}$ is the correlation between the error vectors of configurations $s$ and $c$ from the development stage, and $\lambda \in [0,1]$ trades off relevance (high $\hat{t}_c$) against diversity (low $\max_{s \in S} R_{s,c}$). The candidate with the largest MMR score is added to $S$, and we repeat until $|S| = k$.

\subsubsection{Efficient final selection via successive halving}

{Even with a reduced candidate set, fitting all $k$ configurations on the full training data can be expensive, especially for long time series (Challenge~4 in Section~\ref{sec:rationale}). We also hypothesize that weaker candidates can be identified using a smaller computational budget, for example by fitting them on fewer recent observations, and discarded before committing to evaluation on the full training data. We operationalize this with successive halving, a state-of-the-art bandit-based approach typically used in hyperparameter optimization~\cite{li2018hyperband}: configurations are evaluated on progressively larger windows, and the worst half is eliminated at each round.} We adapt successive halving to ARIMA configuration selection by using progressively larger windows of the target series as the allocated resource and AICc as the evaluation criterion.

The algorithm starts by evaluating all $k$ configurations on a small initial window of the most recent training observations, i.e., those closest to the forecast origin. In each iteration, it discards the worst-performing half of the \emph{currently remaining} configurations based on their AICc scores and doubles the window size for the next round. This process continues until only one configuration remains. The surviving configuration with the best AICc score is selected as the final choice and fit using the complete training data.
Specifically, at iteration $s$, successive halving:
\begin{enumerate}
    \item Uses the most recent $n_s = \min(n_{min} \cdot r^s,\; T-h)$ observations of the training set, where $n_{min}$ is the minimum window size, $r$ is a resource multiplier, and $T-h$ is the training length. When $n_s$ reaches $T-h$, all remaining candidates are evaluated on the complete training data
    \item Evaluates the remaining configurations by fitting ARIMA models on the $n_s$ observations and computing their AICc scores  
    \item Discards the worst-performing half of the currently remaining configurations, keeping the best $\lfloor |\mathcal{S}_s| / \eta \rfloor$ candidates, where $\mathcal{S}_s$ is the active set at iteration $s$ and $\eta$ controls the elimination rate
\end{enumerate}

The AICc scores used in this process serve as a robust model selection criterion, following standard practice in time series model selection (e.g., AutoARIMA). AICc evaluates model fit while penalizing complexity, considering both how well a specification matches the data and its number of parameters~\cite{akaike1998information}. As a relative measure, AICc enables comparison across a set of candidate models, with lower scores indicating better configurations.

Successive halving significantly improves computational efficiency by allocating more resources to promising configurations while quickly discarding poor performers. The combination of meta-learned predictions for initial filtering and successive halving with AICc for final selection provides a robust, data-driven approach to ARIMA model configuration.

\section{Experiments}\label{sec:experiments}

In this section, we describe the experiments conducted to assess the effectiveness of \texttt{MetaARIMA}. The experimental design is structured around the following research questions:
\begin{itemize}
    \item \textbf{RQ1}: Can ARIMA configuration be formulated as an algorithm selection problem, and can metalearning provide an effective and efficient solution?
    \item \textbf{RQ2}: How does a metalearning-based approach for automatic ARIMA configuration perform in comparison to state-of-the-art alternatives and standard classical forecasting methods?
    \item \textbf{RQ3}: How sensitive is the metalearning approach for ARIMA configuration to its individual components, hyperparameters, and to different time series frequencies?
    \item {\textbf{RQ4}: How does \texttt{MetaARIMA} compare with recent zero-shot forecasting TSFMs?}
\end{itemize}

To address these questions, we conduct a series of experiments analyzing various aspects of \texttt{MetaARIMA}'s performance. Our evaluation includes benchmarking against established classical forecasting methods on standard datasets, as well as ablation studies to isolate the impact of specific components. We also analyze how different parameter values in the proposed approach affect its overall performance.

\subsection{Experimental Setup}

\subsubsection{Data}\label{sec:data}

We consider four benchmark datasets from past forecasting competitions: M1~\cite{makridakis1982m1}, M3~\cite{makridakis2000m3}, Tourism~\cite{athanasopoulos2011tourism}, and M4~\cite{makridakis2018m4}. These datasets contain time series with different sampling frequencies. In this study we retain only the monthly, quarterly, and yearly series.

\begin{table}
\caption{Summary of the datasets: number of time series and number of observations}
\label{tab:data}
\begin{tabular}{lll r@{\hskip 0.3cm}r}
\toprule
 & Frequency & Usage & \# time series & \# observations \\
\midrule
\multirow[t]{3}{*}{M4} & Monthly  & \multirow{3}{*}{\rotatebox[origin=c]{0}{Source}} & 48000 & 11246411 \\
 & Quarterly &  & 24000 & 2406108 \\
 & Yearly   &  & 23000 & 858458 \\
 \cmidrule(lr){2-5}
 \addlinespace[4pt]
\multirow[t]{3}{*}{M1} & Monthly  & \multirow{3}{*}{\rotatebox[origin=c]{0}{Target}} & 617 & 44892 \\
 & Quarterly &  & 203 & 8320 \\
 & Yearly   &  & 181 & 4515 \\
 \addlinespace[4pt]
\multirow[t]{3}{*}{M3} & Monthly  & \multirow{3}{*}{\rotatebox[origin=c]{0}{Target}} & 1428 & 167562 \\
 & Quarterly &  & 756 & 37004 \\
 & Yearly   &  & 645 & 18319 \\
 \addlinespace[4pt]
\multirow[t]{3}{*}{Tourism} & Monthly  & \multirow{3}{*}{\rotatebox[origin=c]{0}{Target}} & 366 & 109280 \\
 & Quarterly &  & 427 & 42544 \\
 & Yearly   &  & 518 & 12757 \\
\midrule
Total &  &  & 100.141 & 14.956.170 \\
\bottomrule
\end{tabular}
\end{table}

Table~\ref{tab:data} provides a summary of the datasets. Our analysis encompasses 100.141 time series with over 14 million observations in total. 
In our experimental design, we employ a transfer learning procedure to evaluate the generalization capability of \texttt{MetaARIMA}. Specifically, for each time series frequency (monthly, quarterly, yearly), a separate meta-model is trained exclusively on the corresponding M4 dataset variant (e.g., the monthly meta-model is trained using only monthly series from M4). During inference, this frequency-specific meta-model is then directly applied, without retraining or fine-tuning, to series from the other benchmark datasets (M1, M3, and Tourism) that share the same frequency. This approach enables us to assess the ability of \texttt{MetaARIMA} to transfer knowledge learned from a large, diverse source dataset (M4) to distinct, real-world forecasting problems, mirroring the cross-dataset transfer setting encountered in many practical scenarios where \texttt{MetaARIMA} is expected to be applied.
The forecasting horizon is set to 12 steps for monthly series, 8 steps for quarterly series, and 6 steps for yearly series.

\subsubsection{Methods}

The set of ARIMA configurations $\mathcal{A}$ is obtained by taking the Cartesian product of the parameter ranges in Table~\ref{tab:arima_configs}.

\begin{table}[htb]
\caption{Parameter ranges used to construct the ARIMA configuration space $\mathcal{A}$.}
\label{tab:arima_configs}
\begin{tabular}{ll}
\toprule
Parameter & Values \\
\midrule
$p$ & \{0, 1, 2, 3, 4\} \\
$d$ & \{0, 1\} \\
$q$ & \{0, 1, 2, 3, 4\} \\
$P$ & \{0, 1\} \\
$D$ & \{0, 1\} \\
$Q$ & \{0, 1\} \\
\bottomrule
\end{tabular}
\end{table}

For seasonal series (monthly and quarterly frequencies), this yields $|\mathcal{A}| = 5 \times 2 \times 5 \times 2 \times 2 \times 2 = 400$ configurations. For yearly (non-seasonal) series, we fix $P = D = Q = 0$, giving $|\mathcal{A}| = 50$. All models are fit with the \texttt{statsforecast} package~\cite{garza2022statsforecast}.

On the M4 source data, we run each configuration on every series and record out-of-sample errors (MASE). Features are extracted with \texttt{tsfeatures}~\cite{tsfeatures}, producing $l = 42$ descriptors per series. Errors are binarized at quantile $\tau = 0.25$ (top 25\%) and reduced to $n_{pc} = 50$ principal components via PCA (15 components for yearly series); the correlation matrix $R$ is also computed. A CatBoost multi-target regressor is trained to predict the PCA scores from the features. Hyperparameters of the CatBoost model are tuned by random search (50 iterations) with a 70/30 train/validation split; the best configuration is then refit on all M4 source series.

For a new (target) series, we extract features, predict PCA scores, and invert PCA to obtain configuration scores $\hat{T}_{query}$. MMR with $\lambda = 0.75$ selects $k = 25$ diverse candidates using $\hat{T}_{query}$ and $R$. These are evaluated via successive halving on the training portion of the series: starting with a window of $4m$ observations (where $m$ is the seasonal period), we fit each candidate, score by AICc, discard the worst half, and double the window, repeating until one configuration remains ($\eta = 2$). 

Table~\ref{tab:metaarima_params} summarizes all parameters of \texttt{MetaARIMA} used in the experiments.

\begin{table}[htb]
\caption{Summary of \texttt{MetaARIMA} parameters, their notation, values used in the experiments, and brief descriptions.}
\label{tab:metaarima_params}
\begin{tabular}{llp{7.5cm}}
\toprule
Notation & Value & Description \\
\midrule
\multicolumn{3}{l}{\textit{Development stage}} \\
$\tau$ & 0.25 & Quantile threshold \\
$n_{pc}$ & 50 (15 yearly) & \# principal components\\
$l$ & 42 & \# meta-features \\
\addlinespace[4pt]
\multicolumn{3}{l}{\textit{Inference stage}} \\
$\lambda$ & 0.75 & MMR trade-off \\
$k$ & 25 & \# candidate configurations \\
$n_{min}$ & $4m$ & Initial sample size for successive halving \\
$r$ & 2 & Resource multiplier for successive halving \\
$\eta$ & 2 & Elimination rate for successive halving \\
\bottomrule
\end{tabular}
\end{table}


We compare \texttt{MetaARIMA} with several widely used automatic and fixed-configuration forecasting methods, all implemented using the \texttt{statsforecast} Python package~\cite{garza2022statsforecast}.

\begin{itemize}
    \item \texttt{AutoARIMA}: Automatic ARIMA configuration based on a stepwise search. We use the default \texttt{statsforecast} implementation\footnote{\url{https://nixtlaverse.nixtla.io/statsforecast/docs/models/autoarima.html}} with maximum orders $p,d,q,P,D,Q$ set to 5, 2, 5, 2, 1, 2, respectively, resulting in 94 evaluated configurations. The best configuration is selected according to AICc. 
    \item \texttt{ARIMA}: A fixed ARIMA model with configuration ARIMA(2,1,2), 
    representing second-order AR and MA terms and first-order differencing
    \item \texttt{AR}: A fixed AR model with configuration AR(1) (or ARIMA(1,0,0)), representing first-order autoregression
    \item \texttt{MA}: A fixed MA model with configuration MA(1) (or ARIMA(0,0,1)), representing first-order moving-average
    \item \texttt{ARMA}: A fixed ARMA model with configuration ARIMA(1,0,1)
    \item \texttt{AutoETS}: Automatic selection of ETS (Error, Trend, Seasonality) models according to AICc, using the default \texttt{statsforecast} configuration\footnote{\url{https://nixtlaverse.nixtla.io/statsforecast/docs/models/autoets.html}}.
    \item \texttt{AutoTheta}: The Theta method, automatically configured using a hold-out validation strategy, following the \texttt{statsforecast} implementation\footnote{\url{https://nixtlaverse.nixtla.io/statsforecast/docs/models/autotheta.html}}. This approach evaluates different Theta variants, including fixed versus optimized theta parameters and static versus dynamic models.
    \item \texttt{SeasonalNaive}: A seasonal naive baseline that uses the last observed value in each seasonal cycle as the forecast.
\end{itemize}

\subsubsection{Evaluation}

As described in Section~\ref{sec:data}, we evaluate \texttt{MetaARIMA} in a transfer learning setting. We train the frequency-specific meta-models on the M4 dataset and apply each meta-model, without retraining or fine-tuning, to series from the target datasets (M1, M3, Tourism) that share the same frequency.

The data partitioning protocol for both source and target datasets is conducted as follows. 
In the source dataset (M4), each series is split into three contiguous segments: a training set, a validation set comprising the last $h$ observations before the test set, and a test set comprising the final $h$ observations (unused)\footnote{To our knowledge, the M4 source dataset is independent from the target datasets (M1, M3, and Tourism). Nonetheless, we discard the test segment from each M4 series as a precaution against potential leakage.}. Meta-features are extracted exclusively from the training portion, and ARIMA configurations are fitted on the training set and evaluated on the validation set to obtain performance estimates. In the target datasets (M1, M3, Tourism), each series is split into a training set and a test set (last $h$ observations). Meta-features are extracted from the training portion only, the selected configuration is fitted on the training set, and forecasting accuracy is measured on the held-out test set. This protocol ensures that no future information leaks into the feature extraction or configuration selection stages.

For each method, the selected configuration is fit on the training data and evaluated on the test set under this same protocol.
We use MASE as the evaluation metric~\citep{hyndman2006another,makridakis2018m4}: for monthly and quarterly series, we report the seasonal MASE, and for yearly series, we report the non-seasonal MASE.

\subsection{Exploratory Analysis of ARIMA Configurations}

\begin{figure}[!ht]
    \centering
    \includegraphics[width=\textwidth, trim=0cm 0cm 0cm 0cm, clip=TRUE]{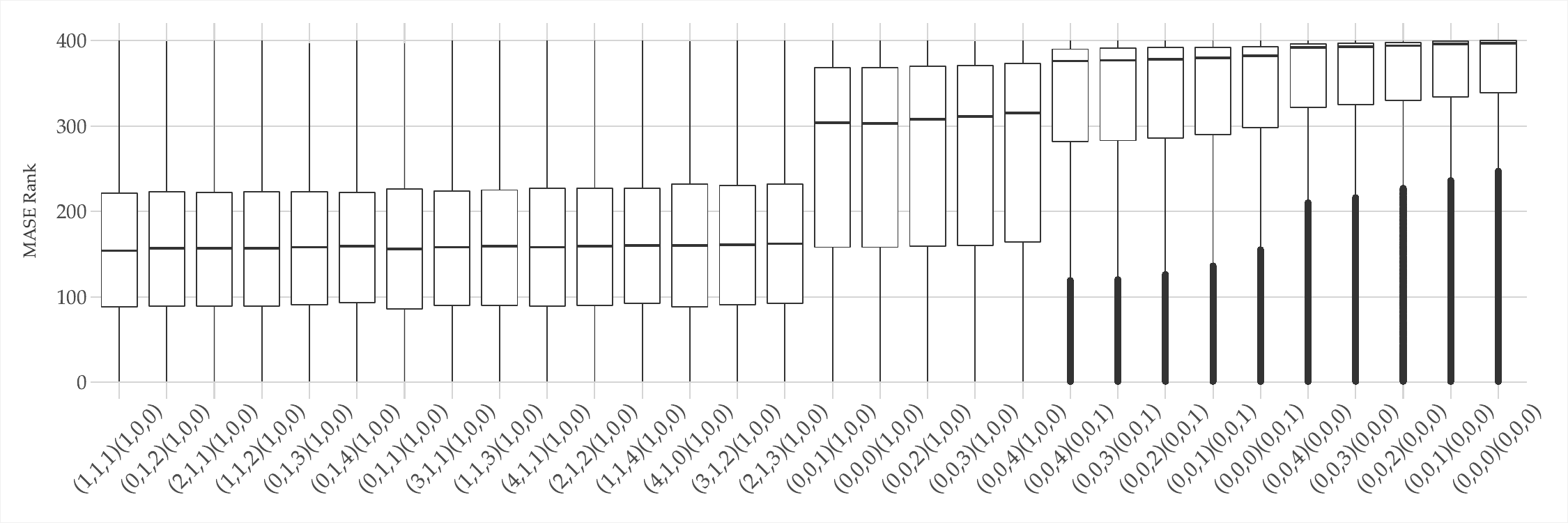}
    \caption{Rank distribution based on MASE of the top 15 and bottom 15 ARIMA configurations across all time series in the datasets. Methods are ordered by average rank.}
    \label{fig:config_rank_dist}
\end{figure}

To better understand the relative performance of different ARIMA configurations, we analyze their performance rankings across datasets. Figure \ref{fig:config_rank_dist} shows the rank distribution based on MASE for the 15 best and 15 worst performing ARIMA configurations across all time series in the monthly datasets (c.f. Table \ref{tab:data}). The configurations are ordered by their average rank, with \texttt{ARIMA(1,1,1)(1,0,0)} achieving the best average rank, followed by \texttt{ARIMA(0,1,2)(1,0,0)}. Conversely, \texttt{ARIMA(0,0,0)(0,0,0)}, which represents white noise, shows the worst average rank performance.

Beyond the performance of specific configurations, a key insight from this analysis is the considerable variability in how configurations perform across different time series. Even the configurations with the best average performance sometimes rank among the worst for certain series, and vice versa. This outcome represents a key motivation for addressing the task of improving the automatic configuration of ARIMA models.

\begin{figure}[!ht]
    \centering
    \includegraphics[width=.7\textwidth, trim=0cm 0cm 0cm 0cm, clip=TRUE]{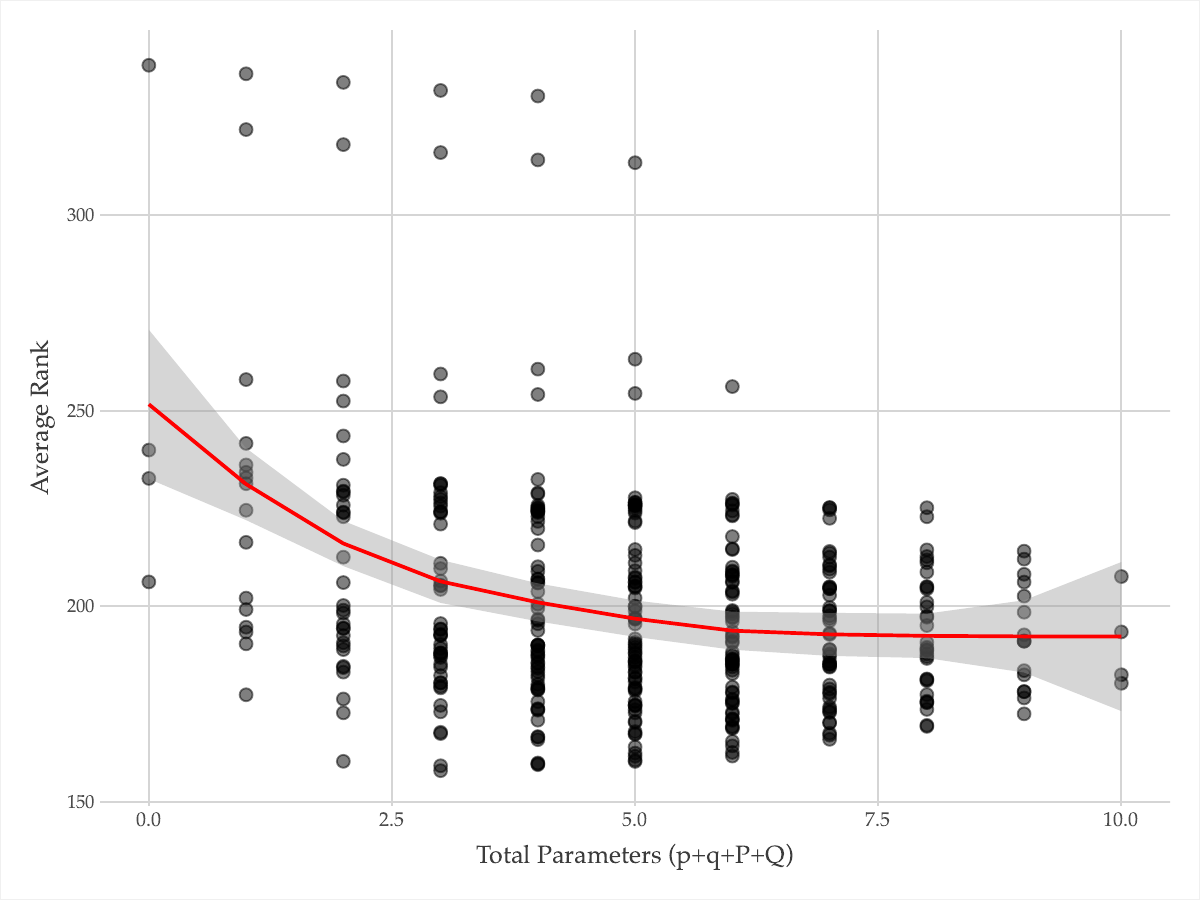}
    \caption{Scatter plot of the average rank of the ARIMA configurations against the total number of parameters. Each point represents a configuration, with the red line representing the LOESS fit.}
    \label{fig:complexity_vs_rank}
\end{figure}

Figure \ref{fig:complexity_vs_rank} shows the scatter plot of the average rank of the ARIMA configurations against the total number of parameters. Each point represents a configuration, with the red line representing the LOESS fit. While there is a slight trend towards better performance with more parameters, the best-performing configurations are not those with the largest number of parameters. 

\subsection{Results}

\subsubsection{Comparing with Classical  Approaches}

\begin{table}
\caption{Average MASE of each approach in each dataset, and respective average rank}
\label{tab:mean_scores_by_ds}
\resizebox{\textwidth}{!}{%
\begin{tabular}{llllllllll}
\toprule
 & \rotatebox{60}{\texttt{MetaARIMA}} & \rotatebox{60}{AutoARIMA} & \rotatebox{60}{AutoETS} & \rotatebox{60}{AutoTheta} & \rotatebox{60}{AR} & \rotatebox{60}{MA} & \rotatebox{60}{ARMA} & \rotatebox{60}{ARIMA} & \rotatebox{60}{SeasonalNaive} \\
\midrule
M1-M & \textbf{0.8709} & 0.916 & \underline{0.9055} & 0.9226 & 1.5173 & 2.1584 & 1.351 & 1.2628 & 1.1233 \\
M1-Q & 1.7027 & 1.6878 & \textbf{1.6609} & \underline{1.6671} & 2.1007 & 4.4207 & 1.9577 & 1.9702 & 2.0775 \\
M1-Y & 3.8196 & \underline{3.7911} & 3.9572 & \textbf{3.6589} & 5.2963 & 9.553 & 4.6296 & 4.2926 & 4.8943 \\
M3-M & \textbf{0.7137} & 0.7367 & \underline{0.7233} & 0.7243 & 1.1869 & 2.1853 & 0.9894 & 0.9792 & 1.0018 \\
M3-Q & \underline{1.1395} & 1.233 & 1.1436 & \textbf{1.1186} & 1.6895 & 3.6976 & 1.5451 & 1.3932 & 1.4253 \\
M3-Y & 3.0899 & 3.1313 & \underline{2.6953} & \textbf{2.6097} & 3.378 & 7.5622 & 3.166 & 3.281 & 3.1717 \\
T-M & \textbf{1.1658} & \underline{1.2165} & 1.2565 & 1.4353 & 4.2227 & 5.0171 & 3.2365 & 2.9088 & 1.3436 \\
T-Q & \textbf{1.4952} & 1.6179 & \underline{1.5989} & 1.6481 & 4.5835 & 6.1959 & 3.7414 & 2.1024 & 1.699 \\
T-Y & \underline{3.4765} & 3.6345 & 3.6299 & \textbf{3.4476} & 4.1316 & 8.1177 & 3.7609 & 3.6421 & 3.7915 \\
\cmidrule(lr){2-10}
Avg. Rank & \textbf{2.0} & 3.22 & \underline{2.44} & \underline{2.44} & 8.0 & 9.0 & 6.22 & 5.67 & 6.0 \\
\bottomrule
\end{tabular}%
}
\end{table}

Table~\ref{tab:mean_scores_by_ds} shows the MASE scores for each approach in each dataset. Bold (underlined) scores represent the best (second best) approach on the corresponding dataset. 
\texttt{MetaARIMA} shows the best or second-best score on 6 out of the 9 datasets, a result that culminates in the best average rank. The average rank quantifies the average relative position (1 for the best approach) across all 9 datasets. 

\texttt{MetaARIMA} is especially effective on seasonal datasets, being the best approach in all monthly variants and also the best or second-best in 2 out of the 3 datasets with quarterly sampling frequency. In non-seasonal datasets, specifically yearly time series datasets for this work, \texttt{MetaARIMA} is outperformed by AutoTheta. However, we note that \texttt{MetaARIMA} still outperforms AutoARIMA in 2 out of the 3 yearly cases.

\texttt{MetaARIMA} outperforms AutoARIMA in 7 out of the 9 datasets.
\texttt{MetaARIMA} also outperforms state-of-the-art exponential smoothing approaches on average rank, namely AutoETS and AutoTheta, an outcome that AutoARIMA fails to achieve. Overall, these results are evidence that \texttt{MetaARIMA} is an effective approach for configuring ARIMA models.

\subsubsection{Ablation and Sensitivity Analysis}\label{sec:sensitivity}

This section presents the ablation and sensitivity analyses used to better understand the performance of \texttt{MetaARIMA}. To control for the number of configurations modeled by \texttt{MetaARIMA}, we focus this analysis on the seasonal datasets (monthly and quarterly variants).

\paragraph{Ablation Analysis}

For the ablation analysis, we systematically removed or modified individual components of \texttt{MetaARIMA}, resulting in the following variants:
\begin{itemize}
    \item \texttt{No-PCA}: \texttt{MetaARIMA} without the PCA step. Here, the targets are binary labels indicating whether a configuration belongs to the top performance quantile, and the meta-learner is a classifier chain of CatBoost classification models (also optimized using random search).
    \item \texttt{No-Binarization}: A variant that removes the top-quantile binarization step and applies PCA directly to the MASE error scores.
    \item \texttt{No-SH}: A variant that does not use successive halving when evaluating the final configuration candidates and instead evaluates them on the full training set.
    \item \texttt{Monte-Carlo}: A variant that replaces successive halving with a Monte Carlo-based evaluation of the final candidates. For each selected configuration, we run 10 simulations; in each simulation, we randomly select a contiguous subset corresponding to 40\% of the training observations and evaluate the configuration on it. The configuration with the best average score across simulations is chosen as the final configuration.
    \item \texttt{Reg-Chain}: A variant that replaces multi-target regression with a regression-chain model to predict the principal components at the meta-level (also optimized using random search).
    \item \texttt{Reg-MO}: A variant that replaces multi-target regression with a multi-output regression model to predict the principal components at the meta-level (also optimized using random search).
\end{itemize}

\begin{figure}[!ht]
    \centering
    \includegraphics[width=\textwidth, trim=0cm 0cm 0cm 0cm, clip=TRUE]{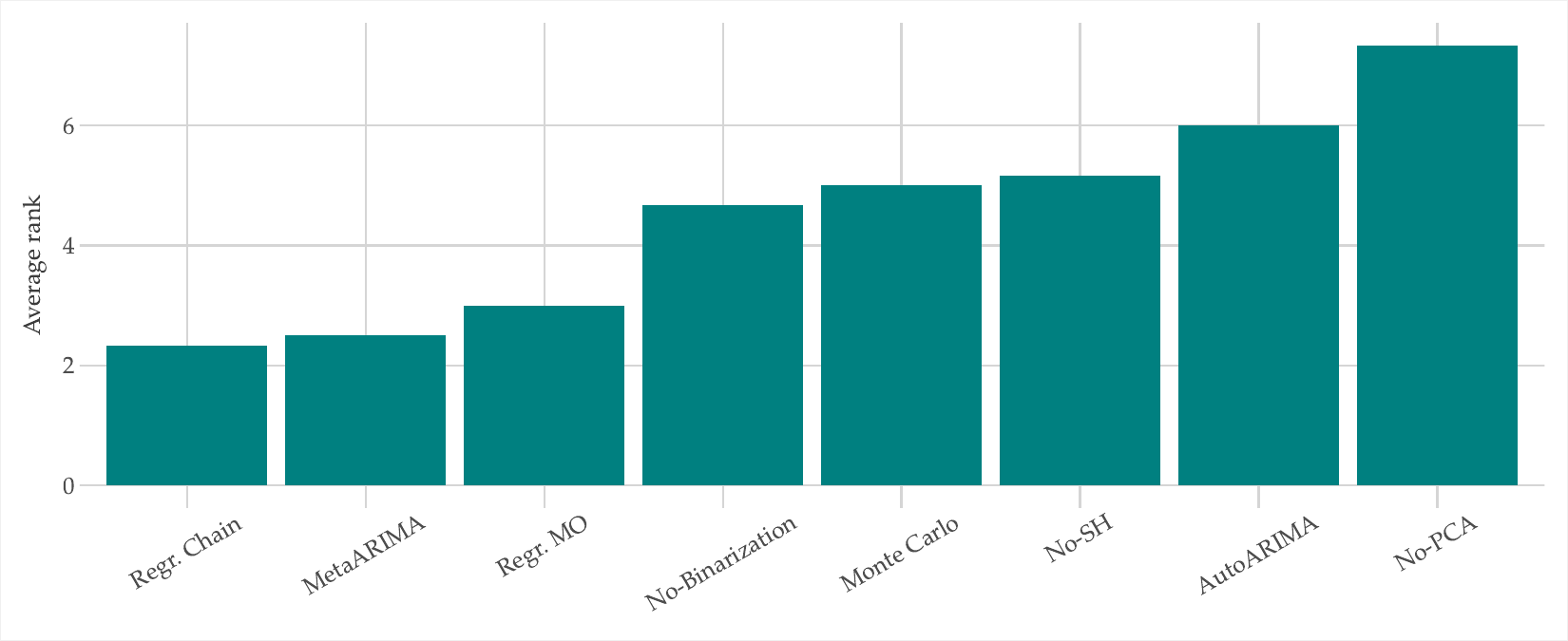}
    \caption{Average rank of different variants of \texttt{MetaARIMA} across all seasonal datasets.}
    \label{fig:ablation_scores}
\end{figure}

Figure~\ref{fig:ablation_scores} shows the average rank performance across all seasonal datasets of the different variants of \texttt{MetaARIMA}. Overall, removing components such as PCA, successive halving, and quantile preprocessing leads to a deterioration in \texttt{MetaARIMA}'s performance. The Monte Carlo variant also performs better than using all data (i.e., No-SH variant), though not as well as the original \texttt{MetaARIMA} with successive halving. Finally, using a regression-chain or multi-output regression model to predict the principal components at the meta-level performs similarly to the original \texttt{MetaARIMA} with multi-target regression. Notwithstanding, the multi-target regression model has the important advantage of being more computationally efficient, as it fits a single model for all targets rather than a separate model for each principal component.

\paragraph{Sensitivity Analysis on the Number of Trials}

\begin{figure}[!ht]
    \centering
    \includegraphics[width=\textwidth, trim=0cm 0cm 0cm 0cm, clip=TRUE]{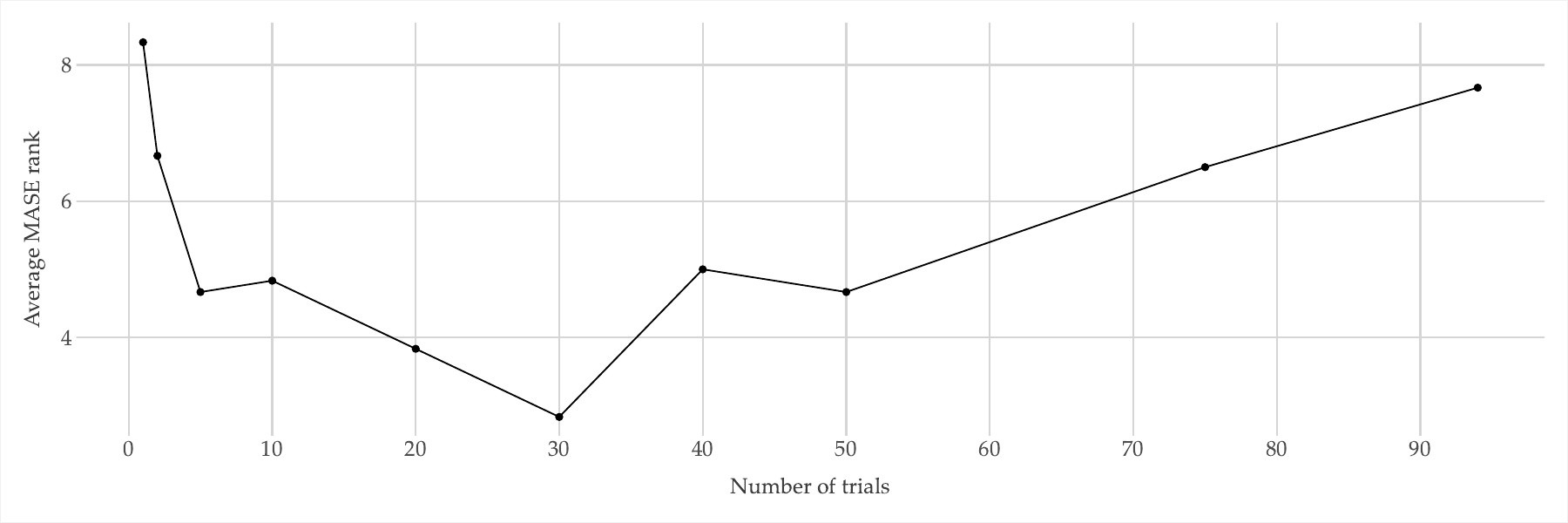}
    \caption{Performance of \texttt{MetaARIMA} across different number of trials. }
    \label{fig:ntrials_scores}
\end{figure}

Figure~\ref{fig:ntrials_scores} shows the performance of \texttt{MetaARIMA} across different number of trials. Figure~\ref{fig:ntrials_scores} shows that \texttt{MetaARIMA}'s performance (lower average MASE rank is better) improves markedly as the number of trials increases from 5 to around 20, reaching its best performance on 30 trials, and then gradually deteriorates as the number of trials continues to grow beyond that, indicating diminishing and eventually negative returns from exploring many more configurations.

This outcome suggests that using a moderate number of trials (around 20) provides the best trade-off for performance and computational efficiency. Increasing the number of trials beyond this point introduces an effect whereby the procedure is likely to select configurations that overfit the validation set, leading to worse generalization performance, on average.

\paragraph{Sensitivity Analysis on $\lambda$}

\begin{figure}[!ht]
    \centering
    \includegraphics[width=\textwidth, trim=0cm 0cm 0cm 0cm, clip=TRUE]{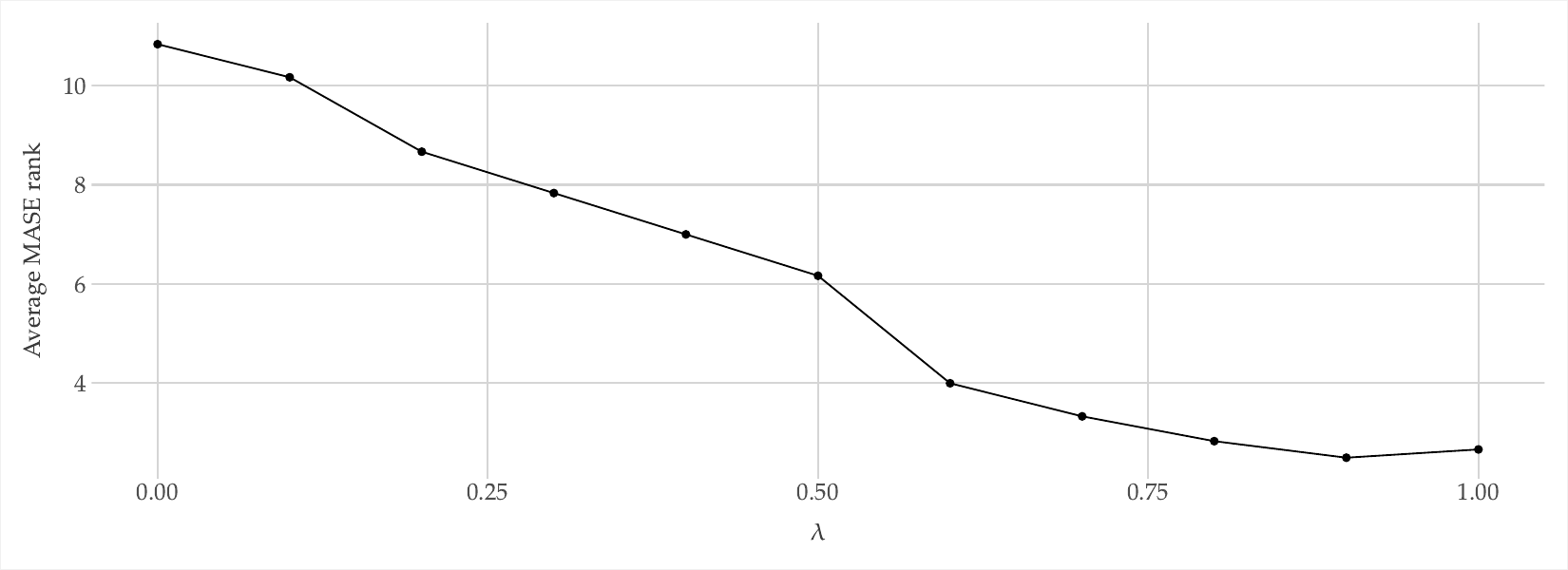}
    \caption{Performance of \texttt{MetaARIMA} across different values for $\lambda$. }
    \label{fig:lambda_scores}
\end{figure}

Figure~\ref{fig:lambda_scores} shows the performance of \texttt{MetaARIMA} across different values for $\lambda$. Recall that $\lambda = 0$ corresponds to selection based purely on diversity, without using the performance prediction scores, whereas $\lambda = 1$ corresponds to selection based only on predicted performance, without explicitly encouraging diversity. The results indicate that performance is very poor when $\lambda = 0$, but improves steadily as $\lambda$ increases, with the average MASE rank dropping up to values around $\lambda \in [0.7, 0.9]$, where performance is best. For values approaching $\lambda = 1.0$, performance deteriorates slightly, although it remains considerably better than for small $\lambda$, suggesting that intermediate values provide the most effective balance between diversity and predicted performance.

\paragraph{Sensitivity Analysis on the Number of Principal Components}

\begin{figure}[!ht]
    \centering
    \includegraphics[width=\textwidth, trim=0cm 0cm 0cm 0cm, clip=TRUE]{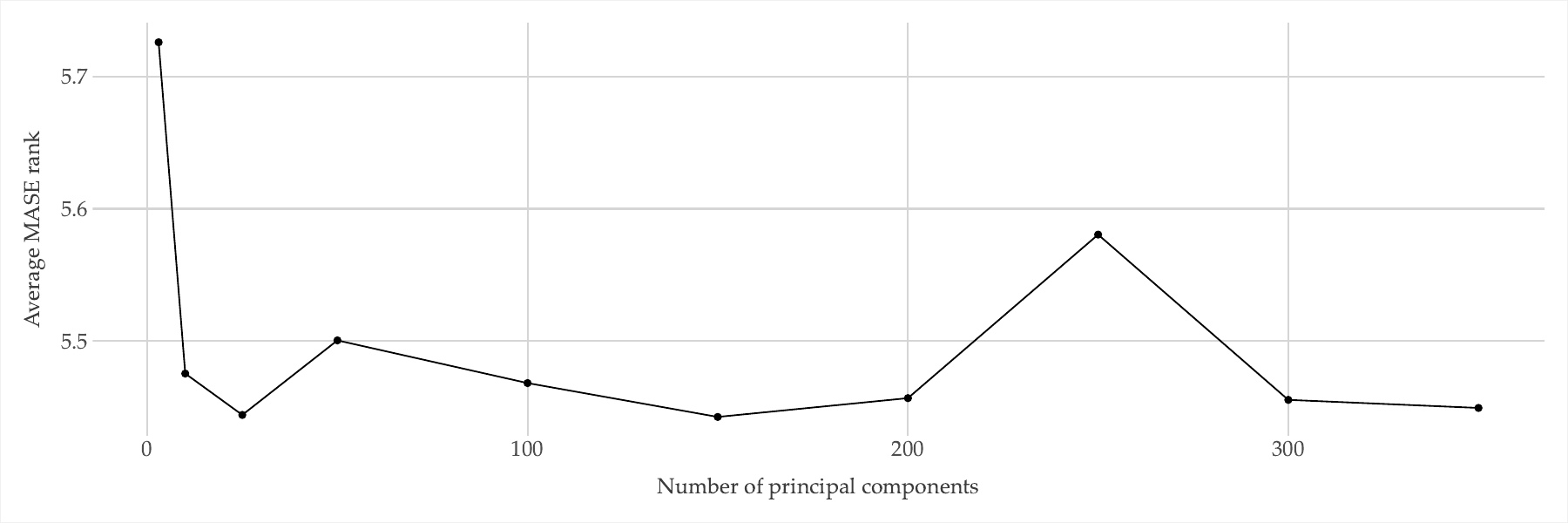}
    \caption{Performance of \texttt{MetaARIMA} across different values for the number of principal components. }
    \label{fig:ncomps_scores}
\end{figure}

Figure~\ref{fig:ncomps_scores} shows the performance of \texttt{MetaARIMA} across different values for the number of principal components, starting from 2 components and going up to 350 (which is close to using all original dimensions). Even with very few components, performance already improves substantially, and as the number of components increases beyond roughly 25--50, performance varies slightly but generally remains strong. We note that our default number of components is 50, a good compromise between performance and computational efficiency.

\paragraph{Sensitivity Analysis on the Quantile Threshold}

\begin{figure}[!ht]
    \centering
    \includegraphics[width=\textwidth, trim=0cm 0cm 0cm 0cm, clip=TRUE]{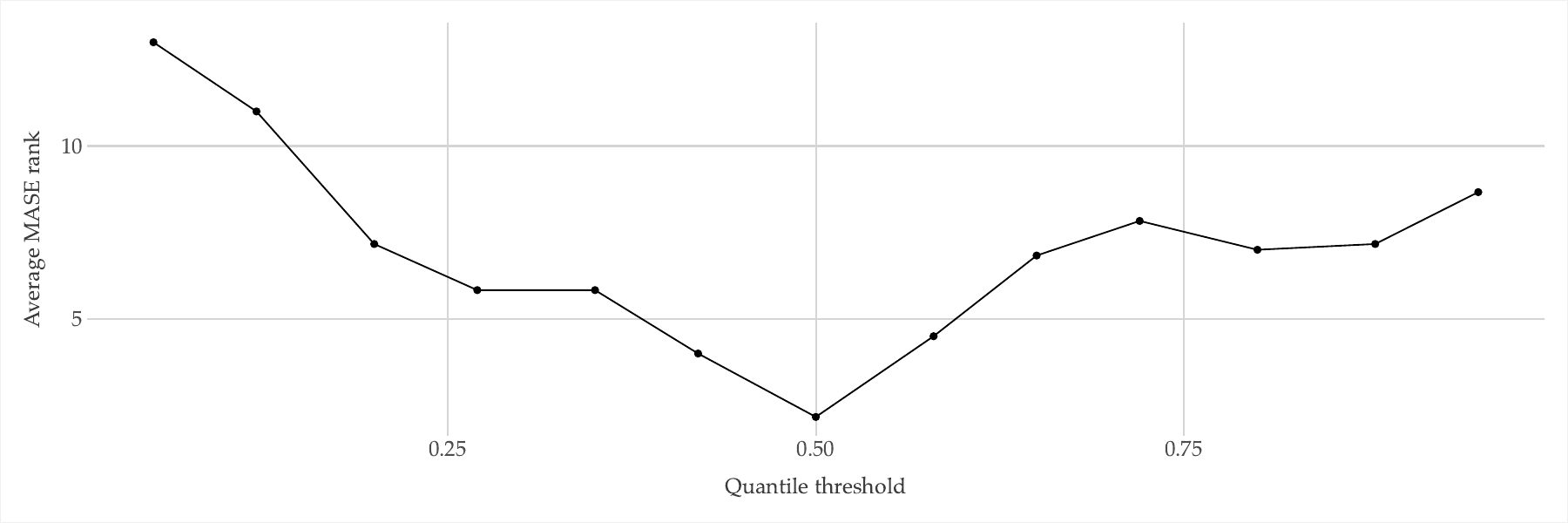}
    \caption{Performance of \texttt{MetaARIMA} across different values for quantile threshold. }
    \label{fig:quantile_scores}
\end{figure}

Figure~\ref{fig:quantile_scores} shows the performance of \texttt{MetaARIMA} across different values of the quantile threshold used to define the top-performing configurations. Very low thresholds (i.e., when only a small fraction of the best configurations is considered top performing) lead to poor performance, but increasing the threshold steadily improves the average MASE rank, with the best results obtained around a threshold of 0.5. Beyond this point, performance gradually deteriorates again as the threshold approaches 1.0, indicating that considering too many configurations in the ``top" set dilutes the signal. Overall, these results suggest that intermediate quantile thresholds provide the most informative supervision signal for \texttt{MetaARIMA}.

\subsubsection{Data Efficiency Analysis}

We compare the data efficiency of \texttt{MetaARIMA} and AutoARIMA by counting the total number of observations each method processes during model selection on the M3-Monthly dataset. For \texttt{MetaARIMA}, this count includes both the observations read during feature extraction (i.e., the full training\ length of each time series, excluding the test set) and those used in the successive halving evaluation of the selected 25 candidate configurations. For AutoARIMA, it includes all observations processed while evaluating the 94 configurations that its stepwise search visits per time series.

Under this accounting, \texttt{MetaARIMA} processes a total of 4,847,310 observations at inference time, whereas AutoARIMA processes 14,140,044 observations. This represents a reduction of approximately 65.7\% in the number of observations required, while \texttt{MetaARIMA} still achieves better forecasting performance.

\subsection{Comparing with TSFMs}\label{sec:comp_tsfm}

{To address RQ4 and contextualise the performance of \texttt{MetaARIMA} against the current state of forecasting research, we compare \texttt{MetaARIMA} with three prominent TSFMs:} Chronos2~\citep{ansari2024chronos}\footnote{\url{https://github.com/amazon-science/chronos-forecasting}}, Moirai2~\citep{liu2025moirai}\footnote{\url{https://github.com/SalesforceAIResearch/uni2ts}}, and TimesFM~\citep{das2024decoder}\footnote{\url{https://huggingface.co/docs/transformers/en/model_doc/timesfm}}. Each TSFM is applied in a direct zero-shot setting (i.e., without any fine-tuning or domain-specific adaptation) following the guidelines provided in their respective repositories. We also include \texttt{AutoARIMA} as a baseline. In the interest of brevity, we restrict this comparison to the monthly dataset variants (M1-M, M3-M, and T-M).

Table~\ref{tab:tsfm} reports the average MASE for each method. Overall, \texttt{Chronos2} achieves the best performance across all three datasets, confirming the strong zero-shot capabilities of state-of-the-art large pretrained forecasters. Nonetheless, \texttt{MetaARIMA} ranks best (tied with \texttt{Chronos2}) or second best in two cases, outperforming both \texttt{Moirai2} and \texttt{TimesFM}, despite relying on a lightweight, interpretable ARIMA model rather than a neural architecture with millions of parameters. This result highlights the effectiveness of the metalearning-based configuration strategy: by selecting well-suited ARIMA specifications, \texttt{MetaARIMA} remains competitive with state-of-the-art deep learning approaches while preserving the transparency and parsimony of classical statistical models.

\begin{table}[th]
\caption{Average MASE of each approach across different monthly datasets}
    \label{tab:tsfm}
    \begin{tabular}{lrrrrr}
    \toprule
     & \texttt{MetaARIMA} & \texttt{AutoARIMA} & \texttt{Chronos2} & \texttt{Moirai2} & \texttt{TimesFM} \\
    \midrule
M1-M & \underline{0.871} & 0.916 & \textbf{0.867} & 0.882 & 0.899 \\
M3-M & 0.714 & 0.737 & \textbf{0.693} & \underline{0.694} & 0.723 \\
T-M & \textbf{1.166} & \underline{1.217} & \textbf{1.166} & 1.25 & 1.359 \\
\bottomrule
\end{tabular}
\end{table}

\begin{figure}[!ht]
    \centering
    \begin{subfigure}[b]{0.48\linewidth}
        \centering
        \includegraphics[width=\linewidth]{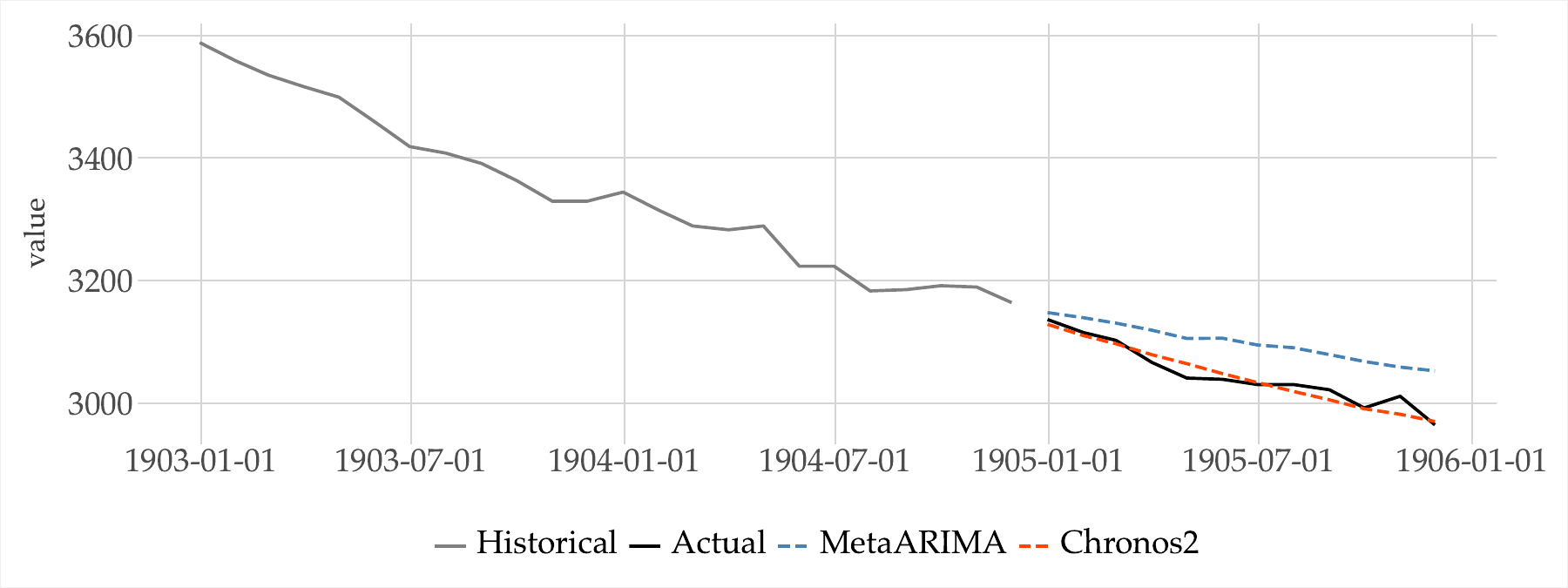}
        \caption{\texttt{Chronos2} achieves the best accuracy.}
        \label{fig:fourpanel-a}
    \end{subfigure}
    \hfill
    \begin{subfigure}[b]{0.48\linewidth}
        \centering
        \includegraphics[width=\linewidth]{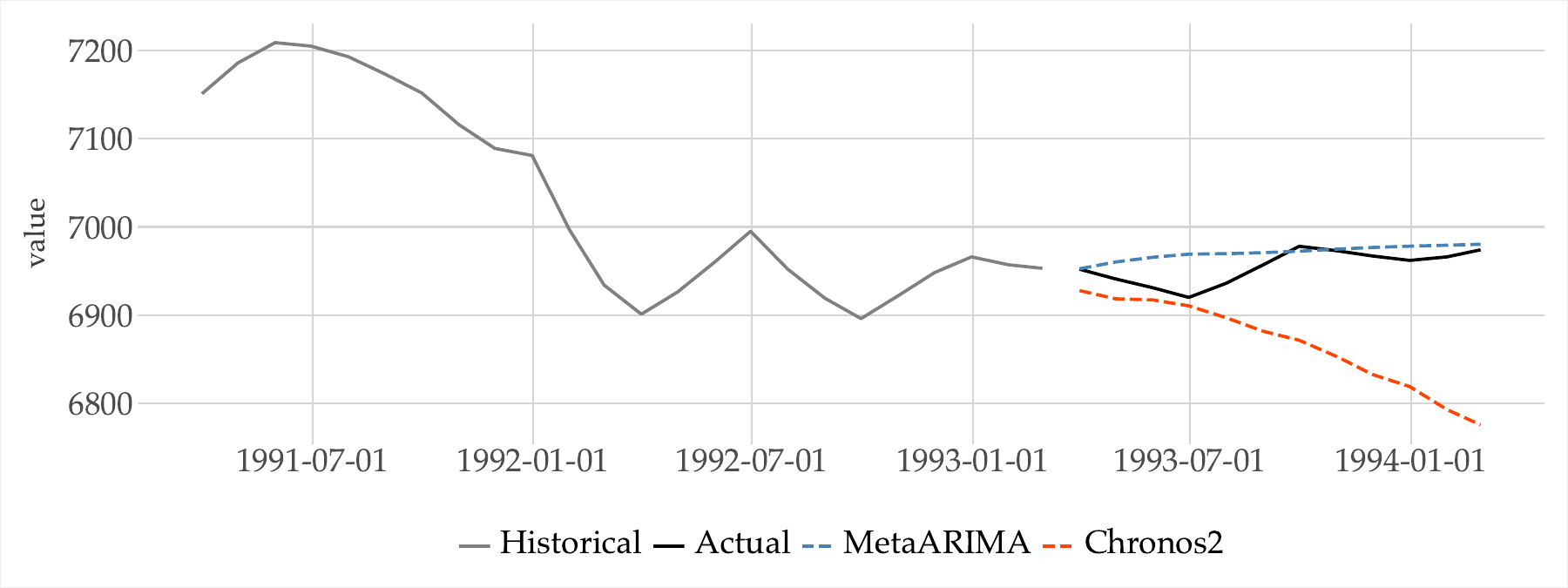}
        \caption{\texttt{MetaARIMA} achieves the best accuracy.}
        \label{fig:fourpanel-b}
    \end{subfigure}
    \\[0.8em]
    \begin{subfigure}[b]{0.48\linewidth}
        \centering
        \includegraphics[width=\linewidth]{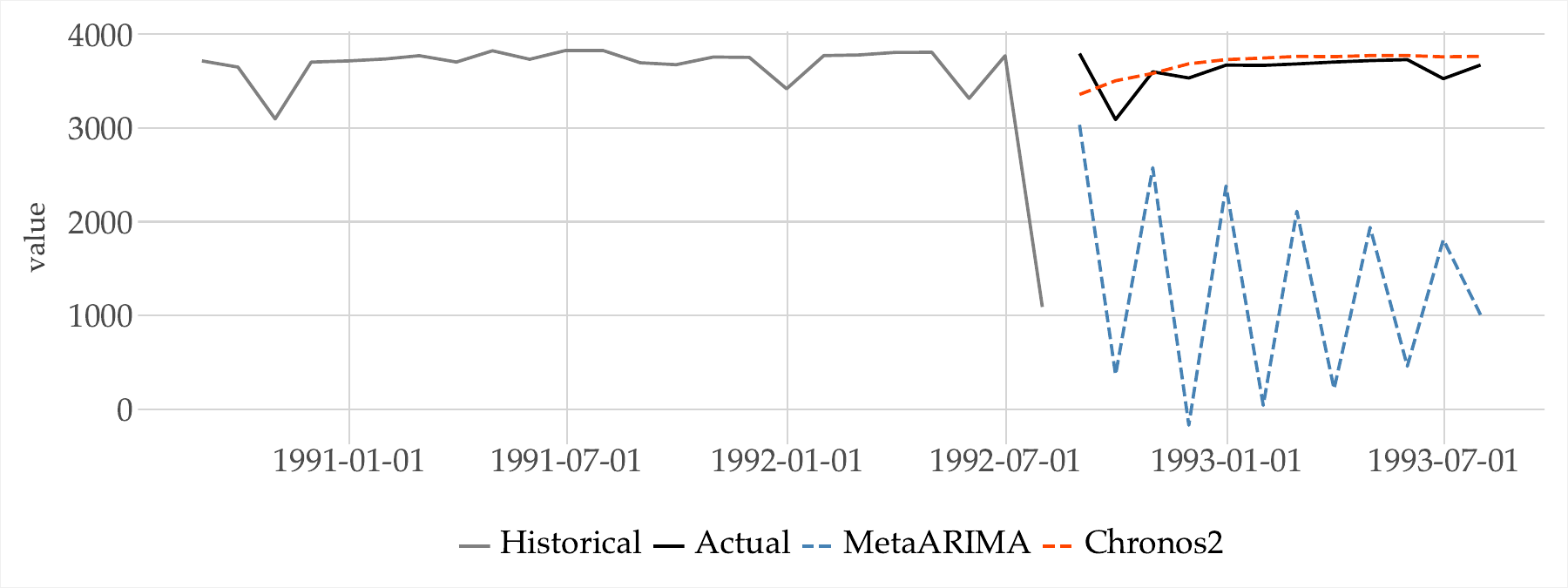}
        \caption{Largest MASE gap in favour of \texttt{Chronos2}.}
        \label{fig:fourpanel-c}
    \end{subfigure}
    \hfill
    \begin{subfigure}[b]{0.48\linewidth}
        \centering
        \includegraphics[width=\linewidth]{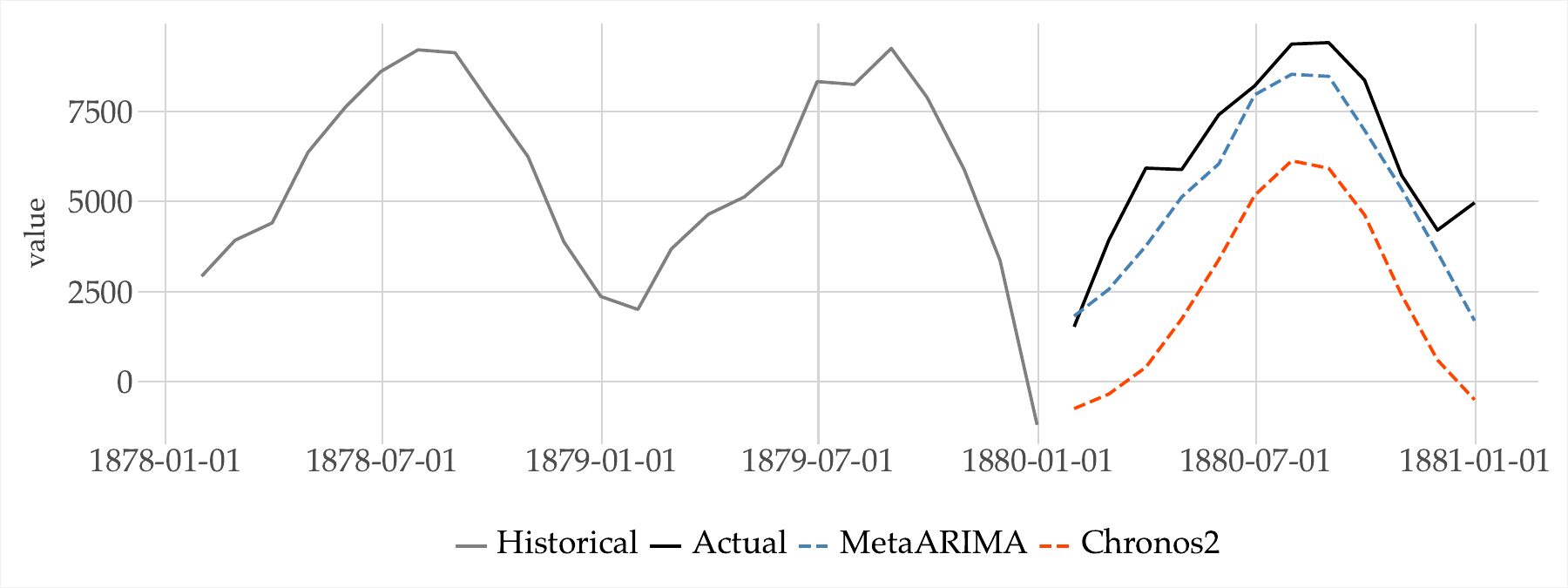}
        \caption{Largest MASE gap in favour of \texttt{MetaARIMA}.}
        \label{fig:fourpanel-d}
    \end{subfigure}
    \caption{Forecasts of \texttt{MetaARIMA} and \texttt{Chronos2} for four time series from the M3-M dataset illustrating contrasting performance scenarios.}
    \label{fig:plots_example}
\end{figure}

To further contextualize the relative performance of \texttt{Chronos2} and \texttt{MetaARIMA}, Figure~\ref{fig:plots_example} presents forecasts for four individual time series from the M3-M dataset, selected to illustrate contrasting outcomes: cases where each method achieves the best accuracy, and cases where the MASE gap between the two is largest in either direction.

\section{Discussion}\label{sec:discussion}

This section presents the answers to the research questions posed in this work and discusses the limitations of \texttt{MetaARIMA}.

\subsection{Research Questions}

The conducted experiments demonstrate that framing the problem of configuring ARIMA as an algorithm selection problem task and applying metalearning is both feasible and effective. By learning from previous configuration performances in historical time series, \texttt{MetaARIMA} can reliably recommend appropriate ARIMA configurations for new time series (RQ1). \texttt{MetaARIMA} also maintains the interpretability and transparency of ARIMA, which are important aspects for many real-world applications. The final selected configuration is fitted to the corresponding time series, following a local training approach typical to ARIMA~\cite{januschowski2020criteria}. 
Yet, \texttt{MetaARIMA} is also, in essence, a global forecasting method, as it learns a mapping from features to top configurations across multiple time series. In effect, \texttt{MetaARIMA} can be regarded as a hybrid strategy that provides a trade-off between global and local methods.

Our results further indicate that \texttt{MetaARIMA} consistently outperforms AutoARIMA and other classical forecasting benchmarks, such as AutoETS and AutoTheta, across a variety of datasets (RQ2). 

We note that the actual raw scores presented in the manuscript are slightly different from the ones reported in the previous works (e.g., original M3 competition) due to the different implementation of the methods. We use the implementation from the \texttt{statsforecast} package for all methods. We remark that our experiments are publicly available in a repository. Regarding the third research question, which concerns the sensitivity of \texttt{MetaARIMA} to its individual components, we observe that \texttt{MetaARIMA}’s performance is influenced by choices such as the $\lambda$ and quantile thresholds. Nevertheless, the method remains robust across a reasonable range of these parameter values (RQ3).

An interesting finding from the sensitivity analyses is that \texttt{MetaARIMA} benefits from promoting diversity among the candidate configurations. The $\lambda$ analysis shows that relying solely on predicted performance ($\lambda = 1$) is suboptimal compared to intermediate values that balance performance and diversity. This supports the hypothesis that ARIMA configurations are often highly correlated in their errors, and selecting a diverse shortlist increases the probability that at least one candidate generalizes well to the target series.

{Regarding RQ4, the comparison with TSFMs (Section~\ref{sec:comp_tsfm}) shows that \texttt{MetaARIMA} outperforms both \texttt{Moirai2} and \texttt{TimesFM} in zero-shot forecasting and remains competitive with \texttt{Chronos2}, the strongest foundation model in our evaluation. 

Conceptually, \texttt{MetaARIMA} shares the transfer learning paradigm of TSFMs in the sense that both learn from a broad collection of time series and generalize to unseen ones. However, \texttt{MetaARIMA} operates at the configuration level rather than the model level, preserving the local estimation properties of classical statistical forecasting.}

This connection to TSFMs also opens an avenue for future work on the representation side. The current implementation of \texttt{MetaARIMA} relies on 42 hand-crafted descriptors from tsfeatures. An alternative would be to augment or replace these features with embeddings produced by a pretrained TSFM, which could capture richer structural patterns in the input series. We deliberately chose tsfeatures in the present work to keep the pipeline lightweight and consistent with the design philosophy of \texttt{MetaARIMA}. Nonetheless, investigating whether TSFM-based embeddings improve configuration accuracy is a promising direction that would create a direct bridge between metalearning-based configuration and foundation model representations.

In this work, we focus on \texttt{tsfeatures} as the input representation for the metalearning stage. However, \texttt{MetaARIMA} is agnostic to the feature extraction framework and alternative time series feature extraction methods could be used. More broadly, one could leverage embeddings produced by a pretrained TSFM as meta-features, potentially capturing richer structural patterns in the input series. Such an integration would create a direct bridge between metalearning-based configuration and foundation model representations.

Beyond our specific architecture, the key takeaway from this work is that there is significant potential to improve the automatic configuration of ARIMA models beyond the current state-of-the-art. We believe such improvements will make ARIMA models more competitive against other approaches, including deep neural networks.

\subsection{Limitations}

While our results demonstrate that \texttt{MetaARIMA} and metalearning approaches show promise for automatically configuring ARIMA models, there is still room for improvement.

\texttt{MetaARIMA} requires historical experiments and performance scores from previous time series, which can be regarded as an impractical computational burden. While this is a limitation, our findings indicate that \texttt{MetaARIMA} achieves greater data efficiency than AutoARIMA while successfully transferring its performance across different datasets and sampling frequencies. In effect, this initial computational investment can be offset by the long-term benefits\footnote{A trained instance of \texttt{MetaARIMA} is provided in the repository and in the \texttt{metaforecast} Python package: \url{https://github.com/vcerqueira/metaforecast}}.

Regarding yearly (non-seasonal) time series, \texttt{MetaARIMA} is outperformed by AutoETS and AutoTheta. A plausible explanation is that the reduced configuration space ($|\mathcal{A}| = 50$) limits the potential gains from metalearning, while the shorter series typical of yearly datasets provide less signal for feature extraction and successive halving. Additionally, exponential smoothing methods may be inherently better suited to capture the patterns prevalent in yearly data as they also outperform AutoARIMA.

More broadly, the transferability of \texttt{MetaARIMA} may depend on the similarity between the source and target domains. In our experiments, two of the target datasets (M1 and M3) share structural characteristics with the M4 source data, while the Tourism dataset is compositionally different; \texttt{MetaARIMA} generalizes well in all three cases. Nevertheless, domains that differ substantially from the source collection, such as financial time series or intermittent demand series, may violate the distributional assumptions learned during training, potentially degrading the quality of configuration recommendations. Retraining or fine-tuning the meta-model on representative source data from the target domain would mitigate this issue.

Another limitation is the number of parameters in \texttt{MetaARIMA}. The method requires setting several parameters, including the number of trials, the number of principal components, the feature extraction method, $\lambda$, other MMR parameters, and the quantile threshold. While our sensitivity analyses demonstrate robustness across reasonable parameter ranges, future work could simplify this parameterization.

\section{Conclusions}\label{sec:conclusions}

We present \texttt{MetaARIMA}, a pretrained configuration policy based on metalearning that brings the transfer learning paradigm to classical statistical forecasting. Rather than pretraining a forecaster, \texttt{MetaARIMA} pretrains a lightweight controller that maps time series characteristics to promising ARIMA specifications, enabling classical models to benefit from historical configuration experiments while preserving interpretability and local estimation. Our experimental results show that \texttt{MetaARIMA} consistently outperforms traditional approaches such as AutoARIMA and AutoETS across diverse benchmark datasets, and remains competitive with recent state-of-the-art TSFMs.

Our findings validate that {configuration-level transfer learning} can effectively improve ARIMA model configuration, achieving better forecasting accuracy while being more data-efficient than existing approaches. \texttt{MetaARIMA} also demonstrates strong generalization, successfully transferring knowledge across datasets.

Beyond our specific architecture, the key takeaway from this work is that there is significant potential for improving the automatic configuration of ARIMA models {by leveraging historical configuration experiments}. We believe such improvements will make ARIMA models more competitive against other approaches, including deep neural networks.

\section*{Acknowledgements}

\small{The authors would like to acknowledge of the anonymous reviewers and the associate editor for their contributions.
This work is funded by national funds through FCT – Foundation for Science and Technology, I.P., within the scope of the research unit UID/00326 - Centre for Informatics and Systems of the University of Coimbra, \url{https://doi.org/10.54499/UID/00326/2025}. This work was also funded by the Agenda “Center for Responsible AI”, nr. C645008882-00000055, investment project nr. 62, financed by the Recovery and Resilience Plan (PRR) and by European Union - NextGeneration EU. Funded by the European Union – NextGenerationEU. Views and opinions expressed are however those of the author(s) only and do not necessarily reflect those of the European Union or the European Commission. Neither the European Union nor the European Commission can be held responsible for them. 
The computational resources used in this study were made available by the Portuguese National Distributed Computing Infrastructure (INCD) through the FCT Advanced Computing Projects 2024.12160.CPCA.A0.}

\vskip 0.2in
\bibliography{bibliography}

\end{document}